\documentclass[]{interact}

\usepackage{epstopdf}
\usepackage[caption=false]{subfig}

\usepackage[numbers,sort&compress]{natbib}
\bibpunct[, ]{[}{]}{,}{n}{,}{,}
\makeatletter
\def\NAT@def@citea{\def\@citea{\NAT@separator}}
\makeatother

\theoremstyle{plain}

\theoremstyle{definition}

\theoremstyle{remark}

\usepackage{epsfig}

\usepackage{amssymb}
\usepackage{amsthm}

\usepackage{graphicx}
\usepackage{amsmath}
\usepackage{amsfonts}
\usepackage{amsmath}
\usepackage{indentfirst}
\usepackage{dsfont}
\usepackage[hidelinks,colorlinks=false]{hyperref}

\newcommand{\la}{\lambda}

\newcommand{\ac}[1]{\left\{{#1}\right\}}
\newcommand{\R}{\mathbb{R}}

\begin{document}


\title{Multiple testing for outlier detection in functional data}

\author{
\name{C. Barreyre\textsuperscript{a}\thanks{CONTACT C. Barreyre. Email: clementine.barreyre@airbus.com}, B. Laurent\textsuperscript{b}, J.-M. Loubes\textsuperscript{c}, B. Cabon\textsuperscript{a} and L. Boussouf\textsuperscript{a}}
\affil{\textsuperscript{a}Airbus Defence and Space, Z.I Palays, 31 rue des Cosmonautes, 31400 Toulouse, France}
\affil{\textsuperscript{b}Institut des Math\'ematiques de Toulouse (UMR 5219), INSA Toulouse, Universit\'e de Toulouse, 135 avenue de Rangueil, 31400 Toulouse, France}
\affil{\textsuperscript{c}Institut des Math\'ematiques de Toulouse (UMR 5219), Universit\'e Paul Sabatier,  Universit\'e de Toulouse, 118 route de Narbonne, F-31062 Toulouse Cedex 9, France}
}

\maketitle

\begin{abstract}
We propose a novel procedure for outlier detection in functional data, in a semi-supervised framework. 
As the data is functional, we consider the coefficients obtained after projecting the observations onto orthonormal bases (wavelet, PCA). A multiple testing procedure based on the two-sample test is defined in order to highlight the levels of the coefficients on which the outliers appear as significantly different to the normal data. The selected coefficients are then called features for the outlier detection, on which we compute the Local Outlier Factor to highlight the outliers. This procedure to select the features is applied on simulated data that mimic the behaviour of space telemetries, and compared with existing dimension reduction techniques.

\end{abstract}


\section{Introduction}
\label{intro}

In this paper, we propose a novel procedure for outlier detection in a set of functional data. \\
Detecting outliers has become an increasing challenge in many areas, such as network intrusion detection, fraud detection, medical anomaly detection, and failure detection, as it was described by Chandola \cite{chandola2009anomaly}. 
An outlier is basically a data that is significantly different from the normal behavior. In addition, several anomalies do not necessarily exhibit similar characteristics. Hence, detecting anomalies must be done by defining the normal behavior in the first place. Then, the deviation measured between an individual and the normal behavior gives good indications of anomalousness. \\
However, as noticed in the same paper \cite{chandola2009anomaly}, defining a normal region that encompasses all the possible normal behaviors is sometimes really difficult. Moreover, an anomaly does not appear necessarily on all the explanatory variables, especially when the data is high-dimensional.\\
In the framework of this paper, the normal behavior can be partially learned thanks to a semi-supervised approach. It means that we can isolate a subset of data that do not contain any anomaly and that will be referred as the nominal set. We therefore have two sets of data, a nominal one and another one containing a small proportion of anomalies, that we want to detect.
This approach is adapted when a first set of data has already been analyzed and reviewed manually.\\
Semi-supervised outlier detection was already treated by \cite{vatanen2012semi} and \cite{sillito2008semi} where the normal behaviour is learned thanks to fixed-background and Gaussian mixture models. \\
We want to detect outliers from functional data since this work is motivated by applications to satellites telemetries, which are measurements of thousands of health parameters through time, hence functional data. Those telemetries exhibit daily-periodic patterns, thus can be logically split into days, leading to consider functions defined on intervals.\\
We suppose that we have $n$ days of a telemetry that is regularly sampled on $p$ times each day. We assume that our observations are corrupted by independent and identically distributed (i.i.d.) Gaussian noise. This corresponds to the following model:
\begin{equation} \label{mod1}
X_{i,j} = f_i(t_j) + \varepsilon_{i,j} ,~ i=1...n,~ j=1,...,p, 
\end{equation}
where $f_i$ is originally defined on a compact set, that can be modeled, without loss of generality, $f_i : [0,1] \mapsto \mathbb{R}$. The  variables $(\varepsilon_{i,j})_{1\leq i\leq n, 1\leq j\leq p}$ are i.i.d. centered Gaussian variables with variance  $\sigma^2$ which is unknown. Since the telemetries are regularly sampled,  we assume that for all $j =1,...,p$,  $t_j=j/p$. \\
Outliers detection applied to functional data was already treated for instance by Hoffmann \cite{kpca_out}, where the Kernel Principal Component Analysis (KPCA) decomposition is used for detecting novelties in hands digits. Ordo{\~n}ez \cite{ordonez2011detection} applied functional data analysis methods on GPS measurements, and the outlier detection is done thanks to the depth metrics. Ren \cite{ren2017projection} proposed to use projections coupled with high-breakdown mean function estimator to detect outliers.\\
Our approach relies on three main steps.
Firstly, we project the data onto orthonormal bases and collect the coefficients resulting from these projections. This approach is really common to cluster functional data, such as in \cite{pbcc} and \cite{antoniadis2013clustering} where coefficients arising from functional basis are used: Reproducing Kernel Hilbert Space (RKHS), wavelets, Functional Principal Component Analysis (FPCA). One can also use the standard Principal Component Analysis (PCA) on the vectors of observations. We choose to project onto the Haar wavelet basis and the principal component basis.\\
In a second step, we select with a multiple testing procedure the projection coefficients that highlight anomalies. The selected coefficients are called features for anomaly detection.\\
Once the features are selected, we compute in a third step the Local Outlier Factor(LOF) \cite{lof} on these features to detect the abnormal behaviours and we compare our results with some common procedures to select features from observations of functional data. We choose this method because it is well adapted to the data we have.
Other methods could be chosen, as the ones described by Barnett \cite{barnett1994outliers}, such as One-Class SVM \cite{ocsvm}, One-Class Random Forest \cite{ocforest}, or depth methods \cite{febrero2008outlier}, for instance.\\
We then apply our procedure on simulated data that mimic the behaviour of space telemetries.\\
The paper is organized as follows. In the first part, we detail two orthonormal bases that are widely used : a wavelet basis and the principal component basis.
In the next section, we apply two-sample tests on each level of coefficients, and then keep the ones that are rejected after controlling the false discovery rate thanks to the Benjamini-Hochberg procedure \cite{benjamini1995controlling}. We consider the Kolmogorov-Smirnov test, as well as two other  tests built on Wasserstein metrics, defined for instance in \cite{ramdas2017wasserstein}.\\
The last section is dedicated to an application on a simulated telemetry reproducing faithfully the behaviour of a real telemetry, and containing all the different types of anomalies that can be observed. The outlier detection method is done thanks to the Local Outlier Factor (LOF) computation. We show in this section the good performances of our procedure to select the best coefficients for outlier detection. We compare it with more classical dimension reduction methods. We also apply the outlier detection on the raw-data, as it was done in \cite{ocsvm_ts}, to show that this naive method is not powerful in this framework.

\section{Definition of the features}

The observations of functional data are high-dimensional, hence it is important to reduce the dimension to concentrate the information.\\
We recall that our observations  obey to Model (\ref{mod1})
where the variables $(\varepsilon_{i,j})_{1\leq i\leq n, 1\leq j\leq p}$ are i.i.d. centered Gaussian variables with variance  $\sigma^2$ and for all $j$, $t_j =j/p \in[0,1]$. We suppose that the nominal days are the $n_0$ first ones, where $n_0<n$.

There exist many ways to represent functional data in a reduced dimension. Some bases are more adapted to some types of functions. Selecting the best basis is in fact a real challenge that has been widely adressed, see \cite{pbcc}, for example. 

We consider  two types of features, the first ones are obtained from a projection onto an orthonormal basis of $\mathbb{L}^2([0,1])$, the second one correspond to a Principal Components Analysis.

\subsection{Projection onto the Haar basis}\label{ond}

We assume that for all $i=1...n$, $f_i \in \mathbb{L}^2([0,1])$, and we denote by $\langle  . , . \rangle $ the usual scalar product in  $\mathbb{L}^2([0,1])$.
$$ \langle f, g \rangle  = \int_0^1 f(t) g(t) dt .$$ 
The functions $f_i$   can be represented in an orthonormal functional basis of $\mathbb{L}^2([0,1])$. See \cite{ramsay} for more details.  If $(\phi_\la)_{\la \in \mathbb{N}^*}$ is an orthonormal basis in $\mathbb{L}^2([0,1])$,
\begin{equation}
f_i(t) = \sum_{\la=1}^\infty \theta_{i,\la} \phi_\la(t), \: {\rm with} \: \theta_{i,\la} = \langle f_i , \phi_\la \rangle. 
\label{general_form}
\end{equation}
From the observations obeying to Model (\ref{mod1}),  the coefficients $\theta_{i,\la}$ are estimated by their empirical counterparts
\begin{equation} \label{coef:ond}
\hat{\theta}_{i,\la} =\frac{1}{p}\sum_{j=1}^p X_{i,j} \phi_\la(t_j).
\end{equation}

We focus here on a wavelet basis, namely the  Haar basis of $\mathbb{L}^2([0,1])$. Let us first recall its definition. We set $\phi_0 = \mathds{1}_{[0,1]}$  and $\psi = \mathds{1}_{[0, 1/2[} - \mathds{1}_{[1/2,1[}$. For all $l\ge 0$, $k\in \Lambda(l) =  \{0,1,\ldots, 2^l - 1\} $, let $\phi_{l,k}(x) = 2^{l/2} \psi(2^l x - k)$. The functions $(\phi_0, \phi_{l,k}, j\ge 0 , k \in \Lambda(l) )$ form the orthonormal Haar basis of $\mathbb{L}^2([0,1])$.
\\
A function $f\in\mathbb{L}^2([0,1])$ can be represented by its expansion onto the Haar basis :
$$f(t) = \alpha\phi_0(t) + \sum_{l \ge 0}\sum_{k \in \Lambda(l)} \beta_{l,k}\phi_{l,k}(t).$$
See Daubechies \cite{daubechies} for more details and other examples of wavelet bases.

For the sake of simplicity, we assume that $p$ is a power of $2$, namely $p = 2^{J+1}$. Since we only have $p$ regularly spaced observations per curve, we only keep the $p$ first wavelet coefficients, which corresponds to all the coefficients  up to the level $l=J$. We define
\begin{equation} \label{Lambdaond}
\Lambda= \ac{0} \cup \ac{ \la=(l,k), \ 0\leq l\leq J, \ k \in \Lambda(l) },
\end{equation}
and
$$ \ac{ \phi_\la, \la \in \Lambda} = \ac{ \phi_0, \phi_{l,k}, \ 0\leq l\leq J, \ k \in \Lambda(l)}.$$
Hence, our initial set of data from Model \ref{mod1} is represented  by the same number of coefficients 
$(\hat{\theta}_{i,\la})_{\la\in \Lambda} $,  $1\leq i\leq n$.  
We can of course recover exactly  the initial data $(X_{i,j})_{1\leq i\leq n, 1\leq j\leq p}$ from the coefficients.
For the next steps, we need the independence of the random variables $ (\hat{\theta}_{i,\la}, {\la\in \Lambda}, 1\leq i \leq n) $ with respect to $\la$ and to $i$. 
This is indeed the case for the three following properties : 
\begin{itemize}
\item  The Haar basis is  orthonormal with respect to the discrete scalar product :
$$ \forall \la, \la' \in \Lambda,  \frac1p \sum_{j=1}^p  \phi_\la (t_j) \phi_{\la'} (t_j)  = \delta_{\la, \la'},    $$
where $  \delta_{\la, \la'} = 0 $ if $\la \neq \la'$ and $1$ otherwise.
\item  For all $i$, the vector  $\mathbf{X}_i = (X_{i,1}, ... , X_{i,p})'$ is a  Gaussian vector.
\item The vectors $(\mathbf{X}_i, 1 \leq i\leq n)$ are independent.
\end{itemize}

In this case, the feature selection will consist in finding the indexes $\la \in \Lambda $ of interest in the decomposition. 
Let us now introduce the other set of features that we consider. 

\subsection{Principal component basis} \label{PCA}
The principal component analysis (PCA) is a very powerful way to reduce the dimension of the data by finding the vectors that recover the best the variance of the data. As this basis depends on the data, we will apply it on a subset $I_0$ of the data known as nominal. We choose  $I_0$ the set of even indexes $\ac{2 i , 1 \leq i \leq n_0/2}$.
Indeed, in order to get independent  features as previously, we will compute the principal components only  on a subset of $n_0/2$ nominal days and project the remaining data on this orthonormal basis. \\
Let $\bar{\boldsymbol{X}}_{I_0} = \frac{1}{n_0/2}\sum_{i\in I_0} \boldsymbol{{X}}_i$ be the mean of the observations. The PCA finds the eigenvectors of the matrix 
$$ \boldsymbol{\Gamma}_{n_0/2} =  \frac{1}{n_0/2} \sum_{i\in I_0}( \mathbf{{X}}_i -  \bar{\boldsymbol{X}}_{I_0}
) ( \mathbf{{X}}_i -  \bar{\boldsymbol{X}}_{I_0})' .$$
Let $(\mathbf{\Phi}_\la)_{\la=1\ldots p}$ be an orthonormal family  of vectors of $  \mathbb{R}^p$ built with the eigen vectors of $\mathbf{\Gamma}_{n_0/2}$, ordered by decreasing eigenvalues. This family is orthonormal for the scalar product $\langle \cdot , \cdot \rangle_p$ in $\mathbb{R}^p$ defined by :
$$ \forall \mathbf{u},\mathbf{v} \in \mathbb{R}^p, ~ \langle \mathbf{u},\mathbf{v}\rangle_p = \frac{1}{p} \sum_{j=1}^p u_j v_j.$$
Finally, we project the vectors $\mathbf{X}_i$ on the subspace generated by the vectors $(\mathbf{\Phi}_\la)_{\la=1\ldots p}$, in order to get 
$$ \mathbf{{X}}_i = \sum_{\la=1}^p \hat{\theta}_{i ,\la} \mathbf{\Phi}_\la$$
where $ \hat{\theta}_{i,\la} =  \langle \mathbf{{X}}_i , \mathbf{\Phi}_\la \rangle_p = \frac{1}{p}  \sum_{j=1}^p X_{i,j}\mathbf{\Phi}_{\la,j}$. We still have the independence of the  random variables $ (\hat{\theta}_{i,\la}, \la\in \Lambda, i \notin I_0) $ with respect to $\la$ and $i$. This is why we need to split the nominal set in two parts in this case. \\ 
The feature selection will consist in finding the levels $\la$ of interest in this decomposition. \\
Once we have presented the features that we consider, we will now see how to reduce the dimension by selecting a small set of features, in order to have good performances for the outlier detection procedure.

\section{ From approximation coefficients to features selection}

We have represented our initial set of data by the  approximation coefficients. Our aim is now to select a small set of coefficients in order to reduce the dimension before applying an outlier detection procedure. Indeed, if we apply directly the outlier detection procedure on the whole set of coefficients, we obtain weak results since the anomalies are not enough separated from the noisy observations, as we will see in the simulation study. There are several ways to reduce the dimension. If the coefficients are naturally ordered (this is the case for the PCA),  usual procedures  consist in keeping the $d $ (with $d<p $) first coefficients in order to maximize the variance explained by those features. For projections onto wavelet bases, we can consider linear procedures of dimension reduction : they consist in keeping all the coefficients up to some level $l=J_0 <J$.  We can also consider nonlinear wavelet thresholding procedures that consist in keeping the $d$ largest wavelet coefficients. These procedures have been proved to be 
very performant for regression function or density estimation procedures, see  \cite{DJKP} for more details. \\
For our purpose, that is to recover abnormal behaviours in our set of curves, it is not necessarily relevant to 
 keep the first principal components or the first levels in  a wavelet decomposition. Moreover,  the thresholding procedure is not well adapted here since the largest coefficients for one day (a value of $i$) are  not necessarily the largest ones for another day. 
We therefore propose a new procedure, based on multiple testing that is well adapted to the problem at hand.

\subsection{Univariate testing at each level}
For many bases, such as the PCA, the usual dimension reduction can be done by taking the $d$ first components that sum up a given percentage of the variance. It is known that, for the clustering usecases for instance, it is a very good way to reduce the dimension, see \cite{ding2004k} for example. It has already also been applied for anomaly detection usecases, as in  \cite{shyu2003novel}. However, no one can assert that an outlier will appear as different in the first principal components. The same objection also holds for projection onto wavelet bases : the first levels do not necessarily contain interesting information for  outlier detection. \\
This is why we have defined a new way to select the features, based on statistical tests, which is adapted to the detection of outliers. \\
We remind that we have represented for all $ 1\leq i\leq n$, our vector of data $\mathbf{{X}}_i $ by the features $(\hat{\theta}_{i, \la})_{\la \in \Lambda}$ with $\Lambda = \ac{1, \ldots, p}$ for the PCA decomposition and $\Lambda$ is defined by (\ref{Lambdaond}) for the Haar decomposition.\\
For each level $\la \in \Lambda$ we would like to know if the vector $\boldsymbol{\hat{\theta}}_{\cdot \la} = (\hat{\theta}_{1,\la}, \ldots, \hat{\theta}_{n,\la})$ contains relevant information on outliers.  The vector  $\boldsymbol{\hat{\theta}}_{\cdot \la}$  is divided into two parts  :   $\boldsymbol{\hat{\theta}}_{0 \la} = (\hat{\theta}_{1,\la}, \ldots, \hat{\theta}_{n_0,\la})$  corresponding to the set that is known to be nominal and  $\boldsymbol{\hat{\theta}}_{1 \la} = (\hat{\theta}_{n_0+1 ,\la}, \ldots, \hat{\theta}_{n,\la})$ for the other values. 
Denote $n_0'$ the number of individuals we keep in the nominal set. If the basis is the Haar wavelet basis, we have $n_0' = n_0$ : all the nominal individuals can be used since the basis is fixed, therefore the features from both subsets are independent. For the PCA basis, we take $n_0' = n_0/2$ since we do not use the features arising from the data $\boldsymbol{X}_{2i}, i = 1,..., n_0/2$ that  were already used to compute the principal components. \\
Denote $\boldsymbol{\tilde{\theta}}_{0\la} = \boldsymbol{\hat{\theta}}_{0\la}$ in the case of the Haar basis, and $\boldsymbol{\tilde{\theta}}_{0\la}$ is composed by the odd indexes of $\boldsymbol{\hat{\theta}}_{0\la}$ for the PCA. \\
We  assume that the components of the vector $\boldsymbol{\tilde{\theta}}_{0 \la} $ are independent and identically distributed (i.i.d.) with common distribution
function $F_\la^{(0)}$ and that the components of the vector $\boldsymbol{\hat{\theta}}_{1 \la} $ are i.i.d. with common distribution
function $F_\la^{(1)}$. Both sets are independent. We will now propose  testing procedures to test  the null hypothesis  $F_\la^{(0)}= F_\la^{(1)}$.

At first we introduce these tests for a single level $\la$. We will treat the problem of multiple testing in the next section. 
\subsubsection{Two sample tests}

Let us suppose we have two independent vectors $\boldsymbol{X}=( X_1, ..., X_{n_0})$ i.i.d. with common   continuous cumulative distribution function $F$ and probability distribution $\mathbb{P}$, and let  $\boldsymbol{Y}=(Y_1, ..., Y_{n_1})$ i.i.d. with common  continuous cumulative distribution function $G$ probability distribution $\mathbb{Q}$.
The generalized inverse functions $F^{-1}$ and $G^{-1}$ are the also called the quantile functions.
 $F$ and $G$ are estimated  by the empirical distribution functions $F_{n_0}$ and $G_{n_1}$, where $\forall t \in \mathbb{R}$, 
$$ F_{n_0}(t) = \frac{1}{n_0}\sum_{i=1}^{n_0} \mathds{1}_{X_i\le t},$$
$$ G_{n_1}(t) = \frac{1}{n_1}\sum_{i=1}^{n_1} \mathds{1}_{Y_i\le t}.$$
Let $X_{(1)} \leq  \ldots \leq X_{(n_0)}$ be the ordered vector $X$.
The quantile function $F^{-1}$ is estimated by  $F^{-1}_{n_0}$ defined as : 
$$
  F^{-1}_{n_0}(p) = 
      \begin{cases}
         X_{(1)}\hspace{1cm}&\text{if } p < 1/{n_0}\\
         X_{(i)}\hspace{1cm}&\text{if } p\in \Big[ \frac{i-1}{n_0},\frac{i}{n_0} \Big[ \text{ and } 2\le i\le n_0\\
          X_{({n_0})}\hspace{1cm}&\text{if } p =1\\
        \end{cases}  
$$
and $G_{n_1}^{-1}$ is computed in the same way. We would like to test the following hypothesis :
\begin{equation}
  \left\{
      \begin{aligned}
        H_0 : F=G \\
        H_1 : F\ne G\\
        \end{aligned}
    \right.
\end{equation}
Many papers deal with the two-sample problem when no prior knowledge is assumed for the shape of the distributions. In this case non parametric tests are used to asses the veracity of the null assumption.  The Kolmogorov-Smirnov test is a reference for this problem  but other  more recent tests can be also implemented built using distance that preserve the shape of the distributions. Here we use  tests  based on the Wasserstein metrics, reviewed for instance in~\cite{ramdas2017wasserstein}. 
\subsubsection{Kolmogorov-Smirnov test}
The Kolmogorov-Smirnov test relies on the fact that under the null hypothesis $H_0$, the distribution of the statistics
$$ D_{{n_0},{n_1}} = \sqrt{\frac{{n_0}{n_1}}{{n_0} + {n_1}}}\sup_{x \in \mathbb{R}}|F_{n_0}(x) - G_{n_1}(x)| $$
does not depend on $F$. 
The test is  rejected when $D_{{n_0},{n_1}} > c_{1-\alpha}$, where $c_{1-\alpha}$ is the defined as the $1-\alpha$ quantile of $ D_{{n_0},{n_1}} $ under $H_0$,  
in order to obtain a  level $\alpha$ for the test.
See \cite{wilcox2005kolmogorov} for further explanation.

\subsubsection{Tests based on  Wasserstein  distances }
When testing the equality of distributions, the choice of the distance used to evaluate the statistical gap between the two samples is important. In the following we introduce a test based on Wasserstein distance.
 First, for $d\geq 1$,  consider  the set 
$\mathcal{W}_q\left( \R^d\right)$ of probabilities with finite $r$-th moment. For $\mu$ and $\nu$ in $\mathcal{W}_q\left( \R^d \right)$, 
we denote by $\Pi(\mu, \nu)$ the set of all
probability measures $\pi$ over the product set $\R^d \times\R^d $
with first (resp. second) marginal $\mu$ (resp. $\nu$). 
The $L_q$ transportation cost between these two measures   is defined as
\begin{equation*}
\label{eq:infwasser}
 {W}_q^q(\mu, \nu) = \inf_{ \pi \in \Pi(\mu, \nu)} \int \left\Vert x - y\right\Vert ^q d \pi(x,y).
\end{equation*}
This transportation cost allows to endow the set $\mathcal{W}_q\left(\R^d\right)$
with the metric $W_r(\mu, \nu)$.
More details on Wasserstein distances and their links with optimal transport problems can be found in  
\cite{rachev} or \cite{villani2009optimal} for instance.

The Wasserstein distance $W_q({P}, {Q})$ between two probability measures ${P}$ and ${Q}$ on $\mathbb{R}$ with $q\ge 1$ finite moments can be easily written as $$ W_q^q({P}, {Q}) = \int_0^1 |F^{-1}(t)-G^{-1}(t)|^q dt$$
where $F^{-1}$ and $G^{-1}$ are the quantile functions of  ${P}$ and  ${Q}$ respectively.
In our framework, we observe two $n$ samples of i.i.d random variables with distribution ${P}$ and  ${Q}$. Let ${P}_n$ and ${Q}_n$ be the empirical distributions, hence the Wasserstein distance between these two empirical distributions is given by
$$ {W}_q^q({P}_n, {Q}_n) =\frac{1}{n} \sum_{i = 1}^n |X_{(i)}- Y_{(i)} |^q. $$
Testing the equality of the two distributions is equivalent to test that the Wasserstein distance ${W}_q^q({P}, {Q})$ is equal to zero.
As a matter of fact, under the assumption that 
$X_1,\ldots,X_n$
are i.i.d. with distribution $P$, $Y_1,\ldots,Y_n$ are i.i.d. with distribution  $Q$ and $P$ and $Q$ have finite $q$-th moment it is easy to conclude that
${W}_q^q(P_n,Q_n)\to {W}_q^q(P,Q)$ almost surely.  However, designing a test requires knowing the asymptotic distribution of a rescaled version of $ {W}_q^q({P}_n, {Q}_n)$ both under $H_0$ to estimate the level of the test and under $H_1$ to evaluate its power. Del Barrio et al. \cite{del1999tests} and references therein give some insights while the case ${P} \neq {Q}$ is tackled in \cite{del2017central}.  \vskip .1in
Yet the asymptotic distribution depends on the distribution ${P}$ which is unknown. Hence we will use the test proposed in \cite{ramdas2017wasserstein} which relies on the following property.
If we consider the image by the distribution function $G$ of the distribution ${P}$ we obtain a distribution ${P}(G^{-1})$ with cumulative distribution function $G\circ F^{-1}$. Under the null assumption $H_0$, then $F=G$ and this distribution is the uniform distribution on $[0,1]$. Hence rather than testing the goodness of fit ${P}={Q}$, we can use this non linear transformation to alternatively test the goodness of fit between the uniform distribution and ${P}(G^{-1})$. The main advantage of this setting is that the asymptotic distribution under the null assumption does not depend on the distribution ${P}$.

Assume that  $f$ and $g$ are the density functions related to $F$ and $G$. Let us suppose that there exists $C\in \mathbb{R}$ such that 
$$\forall t \in \mathbb{R}, ~\frac{g(F^{-1}(t))}{f(F^{-1}(t))}\le C.$$
Let $\gamma = \frac{{n_0}{n_1}}{{n_0}+{n_1}}$. According to Ramdas et al. \cite{ramdas2017wasserstein}, we know that, under the null hypothesis, 
$$ \gamma \times W_2^2(G_{{n_1}\#}{P}_{n_0}, G_{\#}{P})=  \gamma \int_{0}^1(G_{n_1}(F_{n_0}^{-1}(t))- t)^2dt \to \int_{0}^1(\mathbb{B}(t))^2dt$$
and
$$\sqrt{\gamma } \times W_\infty(G_{{n_1}\#}{P}_{n_0}, G_{\#}{P})=  \sqrt{\gamma} \sup_{t\in[0,1]}|G_{n_1}(F_{n_0}^{-1}(t))- t| \to \sup_{t\in[0,1]}|\mathbb{B}(t)|$$
where $\mathbb{B}(t)$ is a Brownian bridge on $[0,1]$.

Consequently, it is possible to build a statistical test, to check the equalities of two distributions thanks to the Wasserstein distance, by using the asymptotic distribution of the test statistics under the null hypothesis to calibrate the quantiles.  
The null hypothesis is rejected if
$$T_{2} = \gamma \int_{0}^1(G_{n_1}(F_{n_0}^{-1}(t))- t)^2dt >c_{2,1 - \alpha}$$
for the 2-Wasserstein test, or if
$$T_{\infty} = \sqrt{\gamma} \sup_{t\in[0,1]}|G_{n_1}(F_{n_0}^{-1}(t))- t| > c_{\infty,1 - \alpha}$$
for the $\infty$-Wasserstein test, where  $c_{2,1 - \alpha}$ is the $1 - \alpha$ quantile of the distribution of $\int_{0}^1(\mathbb{B}(t))^2dt$, and $c_{\infty,1 - \alpha}$ is the $1 - \alpha$ quantile of the distribution of $\sup_{t\in[0,1]}|\mathbb{B}(t)|$.

Since we use the asymptotic quantiles of the test statistics under the null hypothesis, we have carried out some simulations to estimate the level of the test from a non asymptotic point of view.   
\subsection*{Simulation study}
In this example, to evaluate the non asymptotic level of the tests, we  take  ${n_0} = {n_1} = n/2$ and we simulate both samples with standard Gaussian distributions. We simulate $m$ i.i.d. samples: for $k = 1...m$, $X^k \sim \mathcal{N}_{n/2}(0,I_{n/2})$ and $Y^k \sim \mathcal{N}_{n/2}(0,I_{n/2})$, independent of $X^k$.  At each iteration $k$, we test the equality of the distributions of $X^k$ and $Y^k$, from which we get a p-value $p_{k,n}$. As usual, we estimate the level of our tests by the empirical estimator, namely the proportion of tests rejected at a level $\alpha$ among the $m$ Wasserstein tests : 
$$\hat{\alpha}(n) = \frac{1}{m} \sum_{k=1}^m \mathds{1}_{p_{k,n}< \alpha }.$$ 
We repeat it for both Wasserstein tests.
When $n$ is large, we expect $\hat{\alpha}(n)$ to  be close to $ \alpha$.
We choose $ \alpha =0.05$, $m = 5000$ iterations and $n/2$ varies from 50 to 10000. The results  are summarized in Table \ref{table_W2_level}, showing that the  level of the test is close to $5 \%$ even for small sample sizes. This shows that the asymptotic test reaches the desired level very quickly, mostly with the 2-Wasserstein test. \\

\begin{table}
\tbl{Estimated level of the 2-Wasserstein and the $\infty$ - Wasserstein test on Gaussian distributions.}
{
\begin{tabular}{ r | c | c || c | c}
\toprule 
 & \multicolumn{2}{l}{2-Wasserstein test}  & \multicolumn{2}{r}{$\infty$-Wasserstein test}\\ \midrule
  $n$ & $\hat{\alpha}$& $\sqrt{Var(\hat{\alpha})}$ & $\hat{\alpha}$& $\sqrt{Var(\hat{\alpha})}$\\ \\ \midrule
50 & 0.052 & $3.1 \times10^{-3}$ & 0.0394&  $2.8 \times10^{-3}$\\
   100 & 0.046 & $2.9\times 10^{-3}$ & 0.0360& $2.6 \times10^{-3}$\\
   500 & 0.048 & $3.0\times 10^{-3}$ &0.0470 & $3.0 \times10^{-3}$\\
   1000 & 0.045 & $2.9\times 10^{-3}$ & 0.0514& $3.1 \times10^{-3}$\\
   10000 & 0.048 & $3.0\times 10^{-3}$ & 0.0550& $3.2 \times10^{-3}$\\ 
\end{tabular}
}
\label{table_W2_level}
\end{table}

We make the same work with $X^k$ and $Y^k  \sim \mathcal{E}_{n/2}(1)$, and we record the level corresponding to both Wasserstein tests. We present the results  in Table \ref{table_Winf_level}.\\

\begin{table}
\tbl{Estimated level of the 2-Wasserstein and the $\infty$ - Wasserstein test on exponential distributions.}
{
\begin{tabular}{ r | c | c || c | c} 
 \toprule
 & \multicolumn{2}{l}{2-Wasserstein test}  & \multicolumn{2}{r}{$\infty$-Wasserstein test}\\ \midrule
  $n$ & $\hat{\alpha}$& $\sqrt{Var(\hat{\alpha})}$ & $\hat{\alpha}$& $\sqrt{Var(\hat{\alpha})}$\\ \\ \midrule
   50 	& 0.050		& $3.1 \times10^{-3}$	&0.0406 & $2.8 \times10^{-3}$ \\
   100 	& 0.0462	& $3.0 \times10^{-3}$	&0.0400 & $2.8\times 10^{-3}$ \\
   500 	& 0.0462	& $3.0 \times10^{-3}$	&0.0482 & $3.0\times 10^{-3}$ \\
   1000 & 0.0470	&  $3.0 \times10^{-3}$	&0.0536 & $3.2\times 10^{-3}$ \\
   10000 & 0.0456	&  $3.0 \times10^{-3}$	&0.0518 & $3.1\times 10^{-3}$ \\ 
\end{tabular}
}
\label{table_Winf_level}
\end{table}

The level of the test is equivalent with Exponential distributions, even for small values of $n$. 
We have carried out some simulations to compare the performances of the two tests based on $T_2$ and $T_\infty$ with the Kolmogorov-Smirnov test, where the results are presented in the next section.


\subsubsection{Simulation study of the power of the tests}
We simulate independent  samples arising from two different distributions. For $k=1, \ldots, m$,  we simulate 
$$ X^k=(X_1^k , \ldots,  X_{n/2}^k )   \sim \mathcal{N}_{n/2}(0,I_{n/2})$$
$$ Y^k=(Y_1^k , \ldots,  Y_{n/2}^k )   \sim \mathcal{N}_{n/2}(0,I_{n/2})$$
$$ Z^k=(Z_1^k , \ldots,  Z_{n/2}^k )  \sim  \mathcal{N}_{n/2}(\mu,\sigma^2 \times I_{n/2}), $$
 where $(\mu, \sigma^2) \ne (0,1)$.
We denote by $F_{X},F_{Y}$ and $F_Z$ the cumulative distribution functions  of $X, Y$ and $Z$. We know that $F_X = F_Y$ whereas $F_X \ne F_Z$. We denote by $F_{n,X}^k,F_{n,Y}^k$ and $F_{n,Z}^k$ their empirical distribution function for the iteration $k$.\\
At each iteration $k$, we generate  $X^k$, $Y^k$ and $Z^k$ and, based on these observations,  we test the equality of the distribution $F_{X}$ and $F_{Y}$, and then the equality of $F_{X}$ and $F_{Z}$. The first test should be accepted whereas the second should be rejected.\\
Let $p_k^{(0)}, k=1,...,m$ be the p-values corresponding to the test $\ac{F_{X}=F_{Y}}$, and $p_k^{(1)}, k=1,...,m$  be the p-values corresponding to the test $\ac{F_{X}=F_{Z}}$ at the iteration $k$.
Frome these p-values we can compute the true positive rate (TPR) and false positive rate (FPR) for each level $\alpha$ of the test, where $\text{TPR}(\alpha) = \frac{1}{m} \sum_{k=1}^m \mathds{1}_{p^{(0)}_k> \alpha}$ and $\text{FPR}(\alpha) = \frac{1}{m} \sum_{k=1}^m \mathds{1}_{p^{(1)}_k> \alpha}$. The values reported draw a ROC curve from those simulations.\\
We test different values for $n, \mu, \sigma^2$ in order to get the behaviours of each test. We simulate $m=2000$ samples, thus we have $4000$ results of tests in total.\\
\begin{figure}[!t]
\centering
\begin{tabular}{c c }
\includegraphics[scale=0.33]{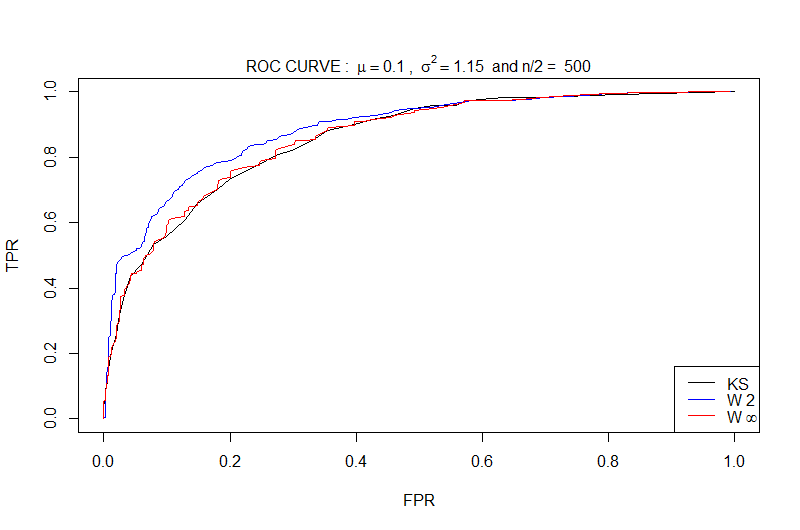} &  \includegraphics[scale=0.33]{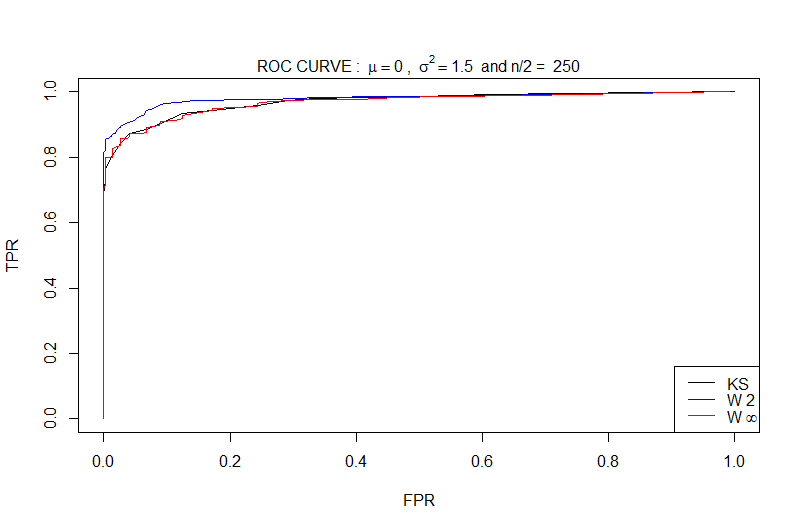}
\end{tabular}
\caption{On the left : ROC curves with $\mu=0.1, \sigma^2=1.15, n= 1000$, on the right : ROC curves, with $\mu=0, \sigma^2=1.5, n=500$}
\label{Normal1}
\end{figure}
\begin{figure}[!t]
\centering
\includegraphics[scale=0.33]{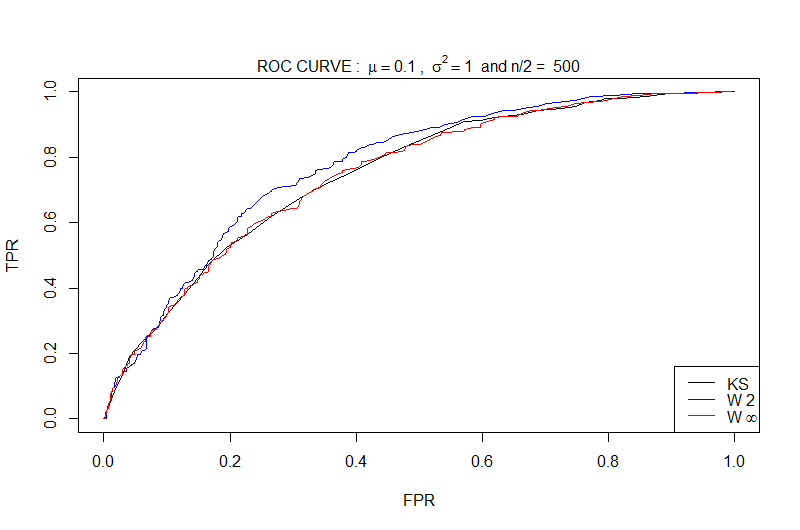}
\caption{ROC curves, with $\mu=0.1, \sigma^2=1, n=1000$}
\label{Normal3}
\end{figure}
As we can see on the figures \ref{Normal1} and \ref{Normal3}, in all the simulations, the 2-Wasserstein test has the best power.

\subsection{Features selection while controlling the false discovery rate}

\subsubsection{A multiple testing framework }

We remind that in both cases described in Sections \ref{ond} and \ref{PCA},  we have defined features   $(\boldsymbol{\hat{\theta}}_{\cdot \la})_{\la \in \Lambda}$  
to represent our data. Each feature $ \boldsymbol{\hat{\theta}}_{\cdot \la}$ is divided in two parts  :   $\boldsymbol{\hat{\theta}}_{0 \la} = (\hat{\theta}_{1,\la}, \ldots, \hat{\theta}_{n_0,\la})$  corresponding to the set that is known to be nominal which are assumed to be i.i.d. with distribution function $F_\la^{(0)}$, estimated from the features $\boldsymbol{\tilde{\theta}}_{0 \la}$, and  $\boldsymbol{\hat{\theta}}_{1 \la} = (\hat{\theta}_{n_0+1 ,\la}, \ldots, \hat{\theta}_{n,\la})$ for the other values, assumed to be i.i.d. with distribution function $F_\la^{(1)}$.  In order to select interesting features, we use the 2-Wasserstein test  to test, for each feature,   the null hypothesis  
$$ H_{0,\la} : \{F_\la^{(0)} = F_\la^{(1)}\}.$$ 
We are therefore dealing with a multiple testing problem since we test in both cases (wavelet or PCA decomposition), $ | \Lambda | = p$ null hypotheses.  
Under $H_{0,\la}$, the p-value $p_\la$ of the test is expected to be (asymptotically) uniformly distributed on $[0,1]$  whereas, under  the alternative,  it is expected to be close to zero. The null hypothesis is rejected at level $\alpha$ for a p-value smaller than $\alpha$. 
Nevertheless, it is well known that, when we deal with many hypotheses, the probability  to have at least one p-value smaller than the level $\alpha$ can be very large. 
Here, for both types of features, we test $ p$ hypotheses and for any $t>0$,
\begin{align*}
\mathbb{P}\Bigg(\bigcup_{\la\in \Lambda} \{p_\lambda < t \}\Bigg) &= 1 - \mathbb{P}\Bigg(\bigcap_{\la\in \Lambda}  \{p_\la > t\}\Bigg)\\
&\sim 1 - (1 - t)^p\\
&  \xrightarrow[p\to \infty]{} 1
\end{align*}
We have used the independence of the random variables $(p_\la, \la \in \Lambda)$.  Indeed, in both cases described in Sections \ref{ond} and \ref{PCA}, the basis $(\phi_\la)_{\la \in \Lambda}$ is orthonormal for the scalar product $\langle \cdot , \cdot \rangle_p$ in $\mathbb{R}^p$. Since the vectors $ \mathbf{{X}}_i $ are Gaussian vectors and since $\hat{\theta}_{i \la} =
\langle \mathbf{{X}}_i  , \phi_\la\rangle_p$ for any $ \la \in \Lambda$, we get that in both cases, the $p$ vectors $(\boldsymbol{\hat{\theta}}_{\cdot \la} )$ for $ {\la \in \Lambda}$ are stochastically independent. \\

For example, we know that for a desired level $\alpha=5 \%$, the probability to reject at least one test among 50 if the null hypotheses are all  true  is already larger than 90\%.
Thus, we will use the procedure proposed by Benjamini and Hochberg \cite{benjamini1995controlling}  to control the false discovery rate. Let us first give some definitions. 
Consider that we have $m$ hypotheses to test where $m_0$ hypotheses are true. $R$ is the total number of rejections. The table \ref{table_example} summarizes the situation.
\begin{table}
\tbl{Multiple testing procedure}
{
\begin{tabular}{  c | c | c | c  }
			
    &  Declared & Declared & Total \\
   & non significant & significant & \\	
   \midrule
   True Null hypotheses &  $U$ & $V$ & $m_0$ \\
   Non-true null hypotheses & $T$ & $S$ & $m-m_0$ \\
   \midrule
   Total & $m-R$ & $R$ & $m$ \\ 
 \end{tabular}
 }
\label{table_example}
\end{table}

In this table, only $R$ and $m$ are known. The false discovery rate (FDR) is defined by 
$$ FDR = \mathbb{E}\Bigg(\frac{V}{max(R,1)}\Bigg).$$
The Benjamini and Hochberg \cite{benjamini1995controlling} procedure allows  to control the FDR.

\subsubsection{Control the FDR with the Benjamini-Hochberg procedure}

Benjamini and Hochberg \cite{benjamini1995controlling} proposed a simple way to control the false discovery rate. Assume  that all the p-values $ (p_k)_{1\leq k\leq m}$ are independent random variables, and let $p_{(1)} \le p_{(2)} \le \ldots \le p_{(m)}$ be the ordered p-values. 
Let $k^*$ be the largest $k$ for which $p_{(k)}\le {k} \alpha/m $, then reject all the hypotheses $H_{0,(k)}, k = 1, ..., k^*$.
Benjamini and Hochberg proved that the FDR for this procedure does not exceed the level $\alpha$.\\
Other procedures exist, and many of them are detailed in \cite{roquain2011type} but we limit ourselves to this procedure since it is very easy to implement.
Note that it is justified in our cases (wavelet or PCA decompositions) since, as explained above, our p-values are independent. This procedure has been chocen since it is really easy to implement, but other procedures can be chosen, where many of them can be found in \cite{roquain2011type}.

\section{Outlier detection with the Local Outlier Factor}
Once  we have selected the features that isolate the best the anomalies in the sense of the comparison of the distributions, we are ready to apply an outlier detection technique on those features. There exists many unsupervised outlier detection methods. Chandola et al. \cite{chandola2009anomaly} detailed many of them. They are often categorized into two categories : distance-based and density-based methods. For instance, the One-Class SVM developed by Sch{\"o}lkopf et al. \cite{ocsvm} is a density-based method that is widely used for outlier detection purposes. 
However, most of time, the telemetry data we are dealing with evolve slowly because of seasonality effects. As a result, there is no immediate separation between the anomalies and the nominal data. Hence, the optimal conditions to use the One-Class SVM are not satisfied.\\
The local outlier factor (LOF) is a score introduced by Breuning et al. \cite{lof} to detect outlier data. In addition of detecting outliers, it returns a score of anomalousness. This method is mixing the density-based and distance-based points of view, and has already been tested on space telemetry data since this method inspired the ESA in the Novelty software \cite{novelty}. It is a local method since this factor depends on how the object is isolated with respect to the surrounding neighbourhood. \\
Suppose we have $n$ objects $x_1, ..., x_n$ to cluster. To simplify the notations, we suppose that for $x_1, x_2, x_3$ all different, $d(x_1, x_2) \neq d(x_1, x_3)$.\\
Let $x \in \ac{x_1, ..., x_n}$ and let $k<n$ be the number of neighbours that we  consider. Choosing the best value for $k$ is not so easy. In \cite{lof}, different values are tested from 10 to 50, and the performance depends on how the data is distributed (different clusters, statistical fluctuations...). As our data depends on seasonnality effects, we opt for considering small values for $k$.\\
- Let $d^k(x)$ be the k-distance of  $x$, which means that for $k$ objects among $x_1, ..., x_n$, the distance to $x$ is closer than $d^k(x)$, and the other $n-k$ points are situated further. Let $N_k(x)$ be the set of  $k$ nearest neighbours of $x$.\\
-  The reachability distance of  $x$ with respect to an object $o$ is then defined as $r_k(x,o) = \max \{d^k(o), d(x,o) \}$. If $x$ and $o$ are sufficiently close, the distance between them is replaced by the k-distance of $o$.\\
-  Then the local reachability density of $x$ is defined as 
$$ lr_k(x) = \frac{k}{\sum_{o\in N_k(x)} r_k(x,o) }.$$ 
-  The local outlier factor is then defined as 
$$ LOF_k(x) = \frac{1}{k}\sum_{o \in N_k(x)} \frac{lr_k(o)}{lr_k(x)}.$$
In other words, the $LOF$ compares the nearest neighbours density distance of a given object with the nearest neighbours density distance of its nearest neighbours. When this score is close to 1, it means that the object is distributed in the same way as its neighbours. When having a large $LOF$, the corresponding object is likely to be an outlier.

\section{Application to simulated data}
\begin{figure}[!t]
\centering
\includegraphics[scale=0.5]{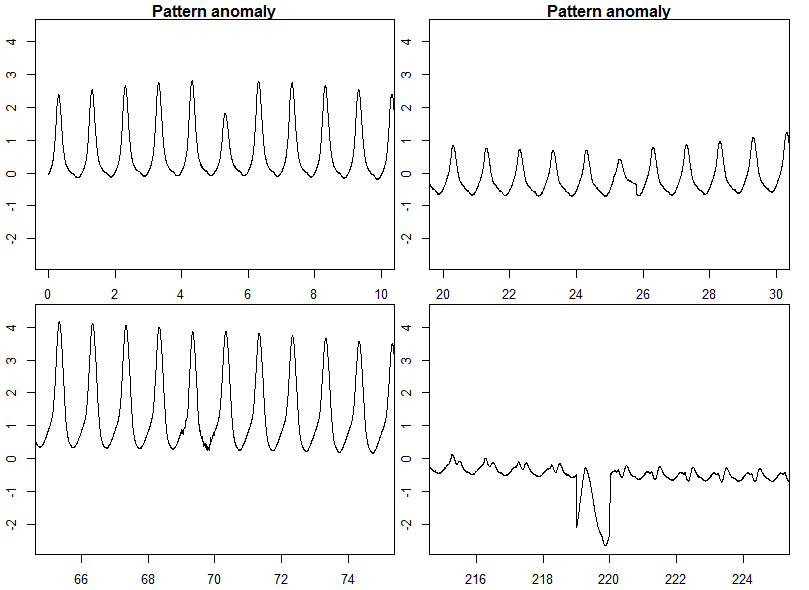}
\caption{Portions of the signal containing the 4 pattern anomalies}
\label{fig_sim1}
\end{figure}

\begin{figure}[!t]
\centering
\includegraphics[scale=0.5]{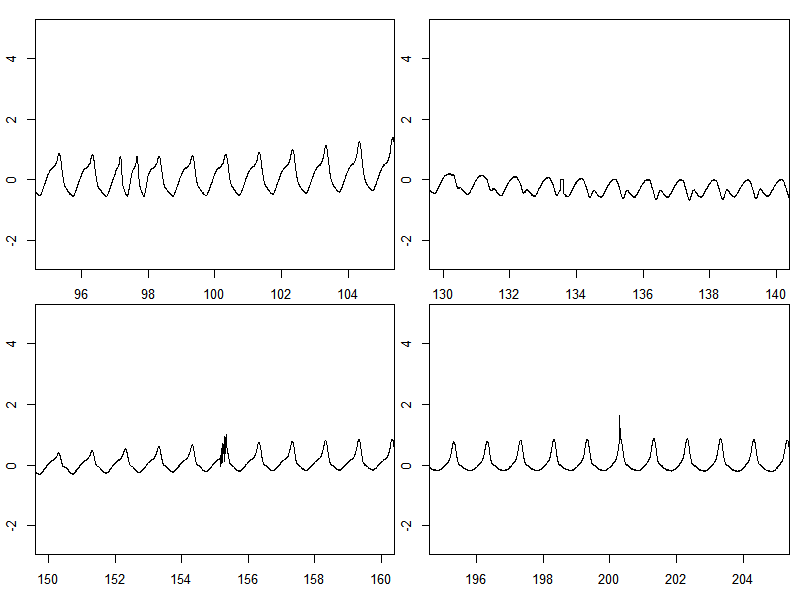}
\caption{Portions of the signal containing the periodicity anomaly (top left ) and the 3 local anomalies}
\label{fig_sim2}
\end{figure}

We apply our outlier detection method  on simulated telemetry that is really close to what we can have on real space telemetries.
This simulation example has been created in order to ease the validation of each method. This example was inspired by geostationary satellites telemetries that have daily periodicity as well as yearly periodicity.\\
We used a combination of periodic signals and given patterns to generate our telemetry. We simulated $n=480$ days of telemetry to get a significant number of signals after splitting the signal into days. We have $p=256$ measurements per day. Each day of telemetry corresponds to an observation. 
The total signal symbolize two year of telemetry, where a year lasts 240 days in this example. We added a Gaussian noise to our observations.\\
We introduced eight anomalies of several types, that represent a complete panel to what can be observed on real telemetries. These anomalies are situated only in the first 240 days. \\ 
The  anomalies that  are introduced are the following: 4 pattern anomalies (change in the pattern or in the amplitude of the data), 3 local anomalies (noise, spikes, data set to default value...) and one periodicity anomaly (two patterns instead of one). The pattern anomalies occur on days 6, 26, 70 and 220, the local anomalies on days 134, 156 and 201, and the periodicity anomaly on day 98.\\
The figures \ref{fig_sim1} and \ref{fig_sim2} show the portions of the signal containing anomalies, where the pattern anomalies are all in Figure \ref{fig_sim1}. Some of the anomalies seem obvious (days $219, 97$), and some other are less pronounced (days $6, 26, 134$). \\ The aim of the study is to retrieve the days with abnormal behaviours. \\
For each day of telemetry, we calculate the features based on the Haar wavelet basis and the PCA basis, and keep the raw-data as a reference result.
We then compute the Local Outlier Factor on the features that are not labeled as nominal. We will test several feature sets: 
\begin{enumerate}
\setcounter{enumi}{-1}
\item The raw data.
\item The full PCA coefficients.
\item The first $d$ coefficients of the PCA representing at least 95\% of the variance.
\item The full Haar wavelet coefficients.
\item The $8$ coefficients from the levels $0,1,2$ of a Haar wavelet basis. 
\item The $8$ coefficients from the third level of a Haar wavelet basis.
\item The $16$ coefficients from the fourth level of a Haar wavelet basis.
\item The PCA components resulting from the feature selection based on the 2-Wasserstein test. 
\item The PCA components resulting from the feature selection based on the $\infty$-Wasserstein test.
\item The wavelet coefficients resulting from the feature selection based on the 2-Wasserstein test. 
\item The wavelet coefficients resulting from the feature selection based on the $\infty$-Wasserstein test.
\end{enumerate}
We do not present the feature selection on Kolmogorov-Smirnov test, because this test is really close to the $\infty$-Wasserstein test, hence returns the same feature selection. 
After selecting our feature sets, we compute the local outlier factor. We have two parameters to calibrate : the number of neighbours to compute the  local outlier factor. and the threshold for the value of the local outlier factor to detect outliers. For the first parameter, we choose  $k=10$ neighbours. As mentionned earlier, $k=10$ seems to be a good choice since the data has a yearly trend, and it is better to consider a number of neighbours that is not too large. \\
For the threshold, we report the days where the LOF is greater than 2, and the ones that are larger than  4.
The results are summarized in the table \ref{table_tnnd}. There are 8 anomalies to detect, and the local anomalies are expected to be harder to spot than the other anomalies.\\
\begin{table}
\tbl{Anomaly found for each feature set}
{
\begin{tabular}{ l | c | l || c | l || l }
Feature &  Anom & False & Anom  & False  & Nb \\
    &  LOF$>$2 & alarm & LOF$>$4 & alarm  & features\\ \midrule
    0 - Raw data		    & 2/8 & 7 & 1/8 & 0 & 256\\
   1 - PCA - full			& 2/8 & 7 & 1/8 & 0 & 256\\
   2 - PCA - 95\%			& 2/8 & 8 & 1/8 & 0 & 3\\
   3 - Haar - full		    & 3/8  & 0 & 1/8 & 0 & 256\\
   4 - Haar (lev 0,1,2)		& 3/8  & 0 & 1/8 & 0 & 8\\
   5 - Haar (lev 3) 		& 3/8 & 0 & 2/8 & 0 & 8\\
   6 - Haar (lev 4) 		& 5/8 & 0 & 3/8 & 0 & 16\\
   7 - PCA  - W2			& 8/8  & 0 & 7/8 & 0 & 17\\
   8 - PCA  - W$\infty$		& 8/8  & 0 & 7/8 & 0 & 14\\
   9 - Haar - W2			&6/8 & 2 & 5/8 & 0 & 4\\
   10 - Haar - W$\infty$	&6/8 & 2 & 5/8 & 0 & 4\\
\end{tabular}
}
\label{table_tnnd}
\end{table}

A first constat is that the Local Outlier Factor computed on the raw-data gives bad results, the same as the ones provides on the full PCA coefficients. There is too many redondant information in the raw-data, and the values are maybe too close to each other in general, hence do not allow to detect the outliers.
This reinforces the fact that projections are really helpful for highlighting outliers. 
\subsection{Comments on the PCA results}
\begin{figure}[!t]
\centering
\includegraphics[scale=0.31]{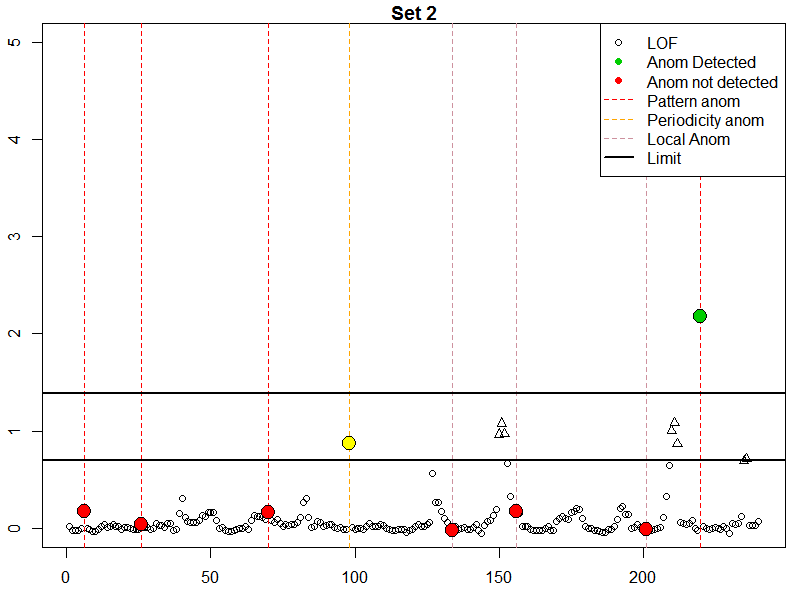}
\includegraphics[scale=0.31]{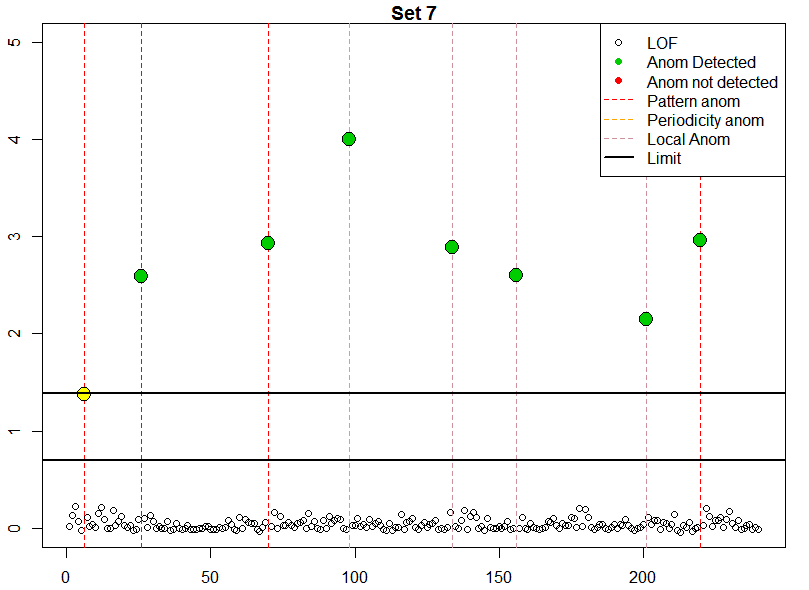}
\caption{LOF for both PCA coefficients selection - common (set2) and novel (set7), and limit (LOF=2,4)}
\label{LOF_PCA}
\end{figure}

At first, one can notice that both two-sample tests generates almost the same feature selection, with, consequently, the same results. 
The set 2, containing the first PCA coefficients, does not contains major information on the outliers. In fact, the 4 first coefficients are not selected when we use the novel feature selection (sets 7 and 8). The selected features that are common from both tests are the ones for which $\lambda \in \ac{10,17,21,29,31,27, 41, 57,58,65,68,69,88,94}$.
It indicates that resuming the full data is not the best way to detect outliers. In fact, the information contained on outliers is unlikely to appear in the first components since the anomaly is rare, thus not representative of a large portion of the variance of the data. The figure \ref{LOF_PCA} shows clearly how performant our procedure is comparatively to the common way to proceed. It enables to isolate better the anomalies to the nominal data. 
Without controlling the FDR, we would have retained 20 features for the 2-Wasserstein test, and 19 for the $\infty$-Wasserstein  test, instead of 17 and 14. As we have chosen $p=256$, it enables to reduce even more the dimension of the data, with better results. In fact, without controlling the FDR, we would have missed one anomaly.\\
An important advantage of our procedure based on the feature selection is that it allows to well separate the values of the LOF of the outliers from the  values of the LOF for nominal date and therefore it is not very sensitive to the threshold : with the value $2$ and $4$ of the threshold, we detect almost the same anomalies for the sets $7,8,9,10$. This is an important property because for the other  cases, the set of detected outliers is very sensitive to the threshold.

\subsection{Comments on wavelet results}
\begin{figure}[!t]
\centering
\includegraphics[scale=0.31]{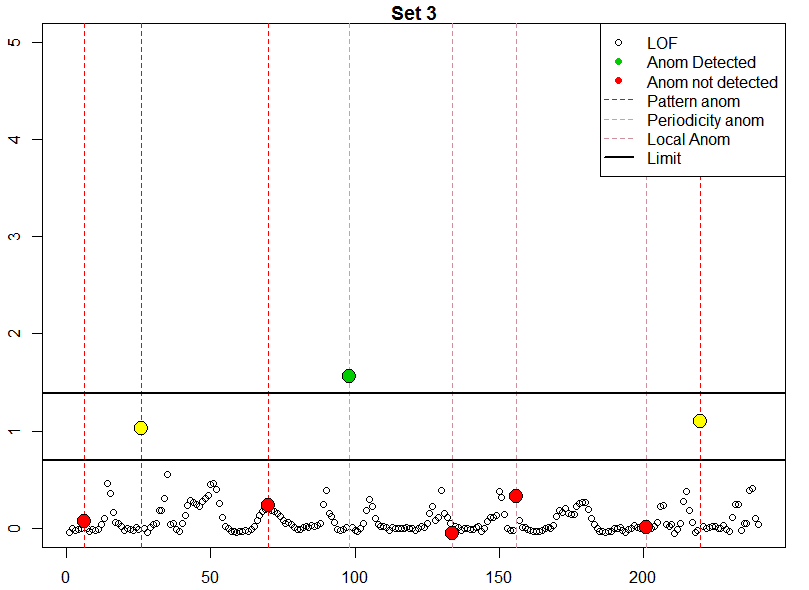}
\includegraphics[scale=0.31]{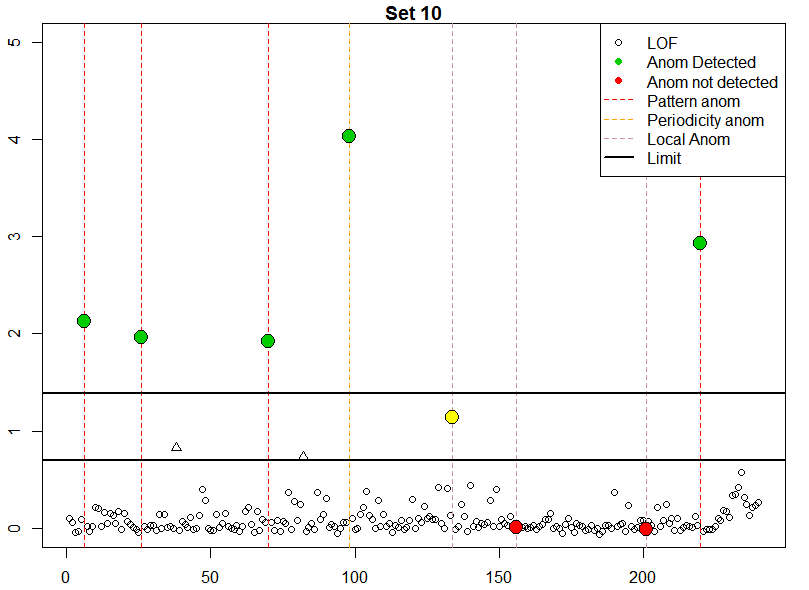}
\caption{LOF for two Wavelet sets coefficients selection -levels where $j \le 2$ (set3), novel procedure (set10), and limit ($LOF>2, LOF>4$)}
\label{LOF_Wavelet}
\end{figure}

We have similar results with the wavelet decomposition. The levels $ l \leq 2$ do not capture any information on the local events. The larger levels, like the level  $l=4$ exhibits easier local events. However, the results are even better thanks to the automatic selection of wavelet coefficients, as we can see on Figure \ref{LOF_Wavelet}. 
Once again, our procedure to select the features enables to isolate better the anomalies to the nominal data. Such results can help to calibrate the level of rejection of the Local outlier factor, since a larger margin will lead to the optimum of the method.
If we look at the 4 wavelet coefficients that were selected after the novel feature selection procedure, with the BH procedure at level $5 \%$, we have: \\
\begin{itemize}
\item 1 position over 4 for $l=2$, 
\item 1 positions over 8 for $l=3$,
\item 1 positions over 16 for $l=4$,
\item 1 positions over 32 for $l=5$,
\end{itemize}
The medium levels are well represented here. Without the BH procedure to control the FDR, we would have retained 12 features, instead of 4. The results are better without controlling the FDR: we would have detected one additional anomaly at the level $LOF = 4$, while removing the false alarms for $LOF>2$.

\section{Conclusion}
This original features selection procedure is really relevant for the detection of outliers.  It reinforces the idea that clustering and outlier detection cannot be addressed by similar methodologies.  
In the case of the PCA, only the first principal components are usually retained for clustering. What we have shown in this paper is that, even if it is the best method to reduce the dimension of the data, it is not the best one to exhibit abnormal behaviours. In fact, those unexpected events do not represent a large portion of the variance of the data, since the outliers are rare and do not have a repetitive signature. Consequently, the anomalies are unlikely to appear clearly as outliers in the very first components, but some further components can get such information. This method is really adapted for the outlier detection in periodical time series, as the space telemetry data.\\
We have also mentioned the case of the Haar wavelet basis. In fact, it is not easy to know which are the levels to retain for such analysis. One can be interested by concentrating the global information, and some other to get details as well. The selection we develop enables to guarantee both types of information on the data.\\
The PCA decomposition with our feature selection procedure gives the best results in this study. The wavelet decomposition combined with the original feature selection procedure gives also good results. An important advantage of the wavelet decomposition  compared with the PCA decomposition is that it is a fixed basis, whereas the PCA is based on a data dependent basis and to keep independence properties, we had to isolate some data to compute the principal components. Moreover, if we want to implement online procedures,  which is often the case when we deal with bigdata sets, a fixed basis is much more relevant. 

Our approach requires to have some knowledge on the data because a set of data that does not contain any anomaly has to be isolated. The difficulty to implement a total unsupervised approach comes from the fact that the distribution of the features for nominal data is unknown.\\
In this direction, a test has been developed by Candelon et al. \cite{candelon2013distribution}, their method is based on bootstrap permutations, which increases a lot the computation time, and it is not suitable for multiple testing, as in our situation.

\bibliographystyle{tfnlm}
\bibliography{biblio_article2}

\end{document}